# Heatmap-Guided Query Transformers for Robust Astrocyte Detection across Immunostains and Resolutions

Xizhe Zhang[1,2]* and Jiayang Zhu[1,2]

[1]Early Intervention Unit, Department of Psychiatry, Affiliated Nanjing Brain Hospital, Nanjing Medical University, Nanjing, China.

[2]School of Biomedical Engineering and Informatics, Nanjing Medical University, Nanjing, Jiangsu, China.

*Corresponding author(s). E-mail(s): zhangxizhe@njmu.edu.cn;

**Abstract**

Astrocytes are critical glial cells whose altered morphology and density are hallmarks of many neuro- logical disorders. However, their intricate branching and stain-dependent variability make automated detection of histological images a highly challenging task. To address these challenges, we propose a hybrid CNN–Transformer detector that combines local feature extraction with global contextual reasoning. A heatmap-guided query mechanism generates spatially grounded anchors for small and faint astrocytes, while a lightweight Transformer module improves discrimination in dense clusters. Evaluated on ALDH1L1 and GFAP stained astrocyte datasets, the model consistently outperformed Faster R-CNN, YOLOv11 and DETR, achieving higher sensitivity with fewer false positives, as confirmed by FROC analysis. These results highlight the potential of hybrid CNN–Transformer architectures for robust astrocyte detection and provide a foundation for advanced computational pathology tools.

# Introduction

Astrocytes are an important type of glial cell in the central nervous system, recognizable by their star-shaped morphology and critical support functions for neurons[1]. They help regulate cerebral blood flow and modulate synaptic activity through the release of gliotransmitters[2]. In particular, astrocytes become reactive in many neurological conditions[3]; trauma, ischemia, infection, and neurodegenerative diseases all trigger astrocyte hypertrophy or other morphological changes[4]. Quantifying such changes in tissue images provides valuable insight into disease mechanisms and progression. Because astrocyte activation often correlates with disease severity and outcomes, an automated method to detect and count astrocytes in histological images would be highly valuable for research and could aid in neuropathological diagnostics[5].

However, it is challenging to reliably identify astrocytes in microscopy images. Astrocytes have intricate, branching structures and often form dense, intertwining networks. This complexity confounds traditional analysis methods: even expert manual counting of GFAP-stained slides (the current gold standard) is laborious and prone to variability, while simple threshold-based algorithms often fail due to inconsistent staining and overlapping cell processes. These issues underscore the need for more robust and fully automated astrocyte detection techniques[6].

Deep learning, especially convolutional neural networks (CNN), has revolutionized biomedical image analysis and now drives state-of-the-art performance in many pathology tasks. Astrocyte detection has begun to benefit from this trend: the first CNN-based astrocyte detector by Suleymanova et al. dramatically outperformed conventional thresholding methods[7], and subsequent deep-learning models have further improved accuracy. However, astrocyte detection still faces unique hurdles. Astrocyte morphology and immunostaining can vary widely across brain regions and imaging protocols, so a model trained on one dataset may falter on another due to domain shifts. Moreover, because astrocytes extend long, entangled processes that overlap with neighboring cells, detecting them reliably requires a larger context than what typical CNN patch-based detectors provide. These challenges indicate that new approaches are needed to handle astrocyte heterogeneity while incorporating global context in the detection process[8].

Early efforts in astrocyte image analysis relied on classical computer-vision techniques such as intensity thresholding and morphological operations[7]. These approaches had limited success, especially in images with high cell density, because they often merged touching astrocytes or required laborious parameter tuning for each dataset. The first deep learning solution for astrocyte detection, *FindMyCells* by Suleymanova et al., proved far more effective[7]. FindMyCells applied a CNN-based object detector to GFAP-stained images and achieved substantially higher accuracy than conventional methods. Building on this foundation, Kayasandik et al. developed a multi-step pipeline combining classical filtering with deep learning to detect astrocytes[9]. Their method used directional filters to identify candidate soma locations before segmenting each cell with a specialized CNN, achieving high accuracy in densely populated images. Another tool, *AICellCounter*, explored a one-shot learning strategy: it allows a cell detector to be trained using only a single annotated image[10]. Although promising for reducing annotation costs, its effectiveness on complex cell types like astrocytes remains to be fully proven.

More recent work has leveraged advancements in deep object detection architectures. Huang et al. adopted the one-stage You Only Look Once (YOLOv5) detector and tailored it for astrocyte detection in both fluorescent and brightfield microscopy images[11]. Their method achieved state-of-the-art accuracy and speed, demonstrating that modern CNN-based detectors can be effectively adapted to glial cells. In parallel, the community has introduced new data resources to support deep learning approaches. Olar et al. curated a large annotated dataset of astrocytes in human brain tissue to facilitate training and evaluation of detection models[12]. Such datasets are expected to drive further improvements by enabling more robust training and providing a standardized benchmark for comparing different methods.

Besides astrocyte-specific work, general cell detection research has informed our approach. Early deep CNN models for mitosis detection in breast cancer histology images showed the power of deep architectures for small object detection[8]. Transformers have recently been applied in computer vision for object detection tasks, with architectures such as Detection Transformer (DETR) demonstrating how self-attention can model global context and predict object locations without heuristic post-processing[13][14].Since the introduction of the end-to-end detection framework DETR, extensive work has focused on improving its convergence speed and small object detection capabilities, including DN-DETR, DAB-DETR, RT-DETRv3, Dynamic DETR and others [15][16][17][18]. These approaches significantly enhance recall for small objects in complex scenes by incorporating dense positive supervision, dynamic query initialization, or multi-scale decoding. DETR and its variants have also shown strong potential in medical images[19][20].

In this paper, we propose a hybrid CNN–Transformer model for automated astrocyte detection that draws inspiration from these advances by combining CNN feature extraction with Transformer-based reasoning. The design combines a CNN backbone for learning local features with a Transformer-based attention module for capturing long-range relationships across the image[13]. By integrating a vision Transformer into the detection pipeline, the model can effectively "see" global context: it links discontinuous astrocytic processes belonging to the same cell and separates overlapping cells in dense networks, overcoming limitations of earlier CNN-only approaches[21]. This hybrid architecture is expected to improve detection accuracy in complex, crowded images while still training with a relatively modest amount of annotated data.

The remainder of the paper is structured as follows. The Materials and Methods section provides a detailed presentation of our proposed hybrid CNN–Transformer framework—including its architectural design and training strategy. In the Results section, we present extensive comparisons between our approach and existing models, along with an analysis of experimental outcomes across multiple datasets and staining conditions. Finally, the Discussion section interprets these findings in the context of current literature, highlights the implications of our work, and concludes with suggestions for future research directions.

# Materials and Methods

## Motivation

Automated astrocyte detection in bright-field IHC is hampered by crowded, low-contrast somata with entangled processes[11]; systematic appearance shifts across stains and scanners [12](e.g., ALDH1L1 vs. GFAP, multiple resolutions); and supervision that marks soma location only coarsely via bounding boxes. Anchor-based CNN detectors, which depend on fixed priors and local receptive fields, tend to miss small or adjacent cells and are brittle under domain shift. Conversely, DETR-style detectors with learned, image-agnostic queries often under-sample subtle somata and converge slowly in dense scenes. Similarly, heatmap-based one-stage detectors like CenterNet, which localize objects as points, struggle to distinguish tightly clustered somata and can produce false posi-tives from background staining due to their reliance on local peak estimation without global context[22].Our design therefore couples a CNN feature pyramid (to capture stain-specific textures and edges) with a lightweight Transformer decoder (to perform global, competitive assignment among candidates), and *grounds* decoder queries in the image by seeding them from a center heatmap. The heatmap focuses computation on plausible somata—especially faint or small ones—while cross-attention over a multi-scale memory mitigates scale uncertainty induced by stain and resolution differences. To align optimization with the task geometry, we adopt focal classification with Hungarian one-to-one matching and L1+CIoU losses, promoting high recall without duplicate predictions in crowded regions[23][24]; Soft-NMS[25] at inference further avoids suppressing near neighbors. Finally, training stain-specific models reduces systematic domain shift while keeping the architecture fixed. The next subsection details the resulting architecture and components.

## Overall Architecture

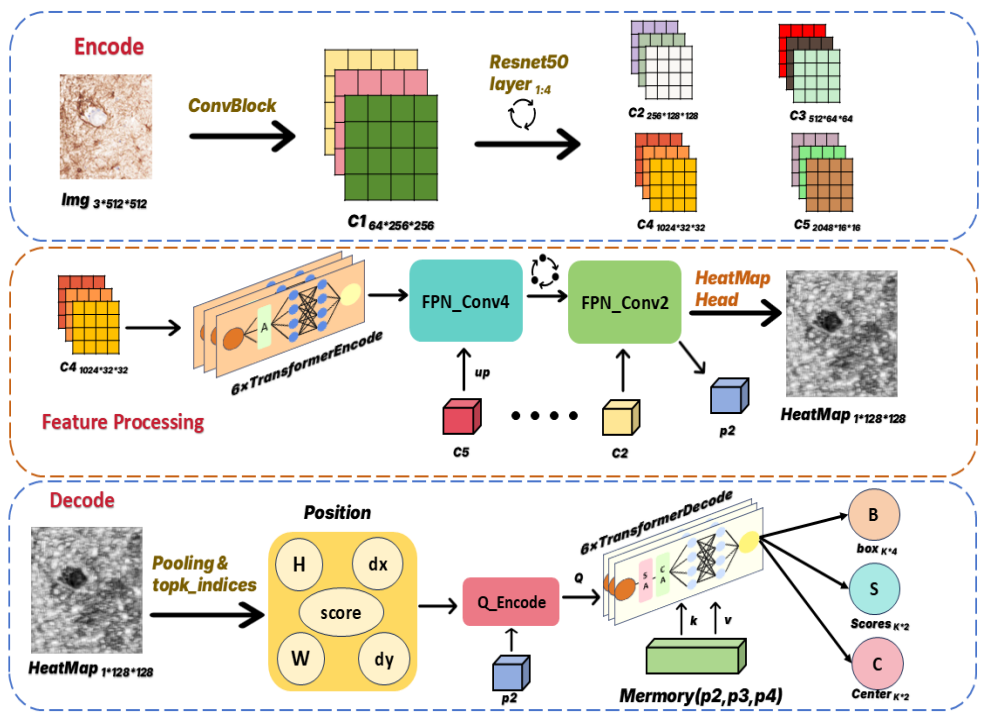

**Fig1. Overall Architecture**

The architecture of our model is a heatmap-guided, transformer-augmented **object detector**. It consists of three main components: (i) a ResNet-50 backbone in which a lightweight Transformer block is inserted at the c4 stage to enhance long-range dependencies, followed by a feature pyramid (FPN) that produces multi-scale feature maps at 1/4, 1/8, and 1/16 resolutions (p2, p3, p4)[26][27]; (ii) a center-heatmap head applied on p2 to generate a single-channel peak map, from which the Top-K local maxima (after 3×3 pool-NMS) are selected as spatially grounded query anchors; (iii) a dynamic query decoder that initializes one query per anchor by bilinearly sampling p2 and concatenating normalized coordinates and positional encodings, and then refines these queries through

L layers of self-attention and cross-attention over a memory built by channel-aligning and concatenating p2, p3, p4. The decoder outputs foreground/background logits and anchor-relative box offsets, which are decoded to (cx, cy, w, h); inference adopts Soft-NMS. Training uses Hungarian matching with a composite loss comprising heatmap focal loss, query-level focal classification loss, L1 regression, and CIoU[28]. Unlike DETR and related detection transformers that rely on a fixed set of learned object queries, our approach derives **data-dependent** queries from the heatmap at 1/4 resolution, improving recall on small and crowded lesions while preserving spatial coherence.

## Hybrid Training Strategy

We propose a heatmap-guided, transformer-augmented object detector. A ResNet-50 backbone extracts multi-level features $\{c2, c3, c4, c5\}$; a lightweight Transformer block is inserted at $c4$ to enhance long-range dependencies, and a feature pyramid produces $\{p2, p3, p4\}$ at 1/4, 1/8, and 1/16 resolution. On $p2$ (1/4), a single-channel *center heatmap* head predicts logits $H$. Local maxima are selected by $3 \times 3$ pool-NMS, and the Top-$K$ peaks ($K = 80$ by default) serve as *data-dependent* query anchors.

*Targets and query initialization.*

For each ground-truth box $j$ with center $(\mu_j^x, \mu_j^y)$, the heatmap target is

$$H^\star(x, y) = \max_j \exp\left(-\frac{(x - \mu_j^x)^2 + (y - \mu_j^y)^2}{2\sigma_j^2}\right),$$

where $\sigma_j$ is a radius derived from the object size. Let $\mathcal{S} = \{(u_i, v_i)\}_{i=1}^K$ be Top-$K$ peak locations on $H$ after pool-NMS. For each $(u_i, v_i)$ we bilinearly sample $p2$ to obtain $f_i$, concatenate normalized coordinates and positional encodings, and project to obtain the initial query

$$q_i^{(0)} = \phi([f_i, \tilde{u}_i, \tilde{v}_i, \text{PE}(u_i, v_i)]) \in \mathbb{R}^d,$$

forming $Q^{(0)} \in \mathbb{R}^{K \times d}$ ($d = 256$).

*Cross-scale memory and decoder.*

We build a memory $M$ by channel-aligning each $p\ell$ with $1 \times 1$ conv, adding 2D positional encodings, and flattening/concatenating:

$$M = [\psi_2(\text{PE}(p2)); \psi_3(\text{PE}(p3)); \psi_4(\text{PE}(p4))] \in \mathbb{R}^{N \times d}.$$

A LiteDecoder with $L$ layers ($L = 6$ by default; $n_{\text{head}} = 8$) iteratively updates queries via

$$Q^{(\ell+1)} = \text{FFN}\left(\text{CrossAttn}(\text{SelfAttn}(Q^{(\ell)}), M)\right),$$
$$\text{where} \quad \ell = 0, \dots, L - 1.$$

Two heads predict foreground/background logits $p_i$ and anchor-relative box offsets $(\Delta x_i, \Delta y_i, \Delta \log w_i, \Delta \log h_i)$.

*Anchor-relative decoding and inference.*

Let $(\hat{c}_x^{(a)}, \hat{c}_y^{(a)})$ denote the anchor (peak) in normalized coordinates. We decode

$$c_x = \hat{c}_x^{(a)} + s_\Delta \tanh(\Delta x), c_y = \hat{c}_y^{(a)} + s_\Delta \tanh(\Delta y), w = w_0 e^{\Delta \log w}, h = h_0 e^{\Delta \log h}$$

with $s_\Delta = 0.3$ and $(w_0, h_0) = (0.08, 0.08)$ by default, yielding $(c_x, c_y, w, h) \in [0,1]$. At inference, boxes are mapped to pixels and *Soft-NMS* is applied[25].

*Training: matching and losses.*

We adopt Hungarian matching between predicted queries and ground truths using a composite cost

$$\text{Cost} = \lambda_{cls}\text{BCE}(p, 1) + \lambda_{L1}\|b - \hat{b}\|_1 + \lambda_{IoU}(1 - \text{CIoU}(b, \hat{b})) + \lambda_{ctr} d_{center}(b, \hat{b}).$$

with default weights $\lambda_{cls} = 3$, $\lambda_{L1} = 5$, $\lambda_{IoU} = 4$, $\lambda_{ctr} = 4$. The total loss is

$$\mathcal{L} = w_{hm}\mathcal{L}_{Focal}(H, H^\star) + w_{cls}\mathcal{L}_{Focal}(y, \hat{y}) + w_{L1}\|b - \hat{b}\|_1 + w_{IoU}(1 - \text{CIoU}(b, \hat{b})),$$

we set default weights $w_{hm} = 2.0$, $w_{cls} = 1.0$, $w_{L1} = 6.0$, $w_{IoU} = 2.0$, and focal parameters with $\alpha = (0.25, 0.75)$ and $\gamma = 1.5$. Unlike DETR, which uses a fixed set of learned queries, our queries are heatmap-derived and spatially grounded, improving recall for small and crowded lesions while preserving spatial coherence.

## Astrocyte Detection for Different Stains

We train two separate models – one for ALDH1L1-stained images and one for GFAP-stained images. This follows the approach of the baseline, given the substantial differences in appearance between ALDH1L1 and GFAP slides [12]. The ALDH1L1 model learns to detect mostly rounded astrocyte cell bodies with homogeneous staining, whereas the GFAP model learns the filamentous patterns and often fragmented appearance of GFAP-positive astrocytes. The same architecture and hyperparameters are used for both, but training them separately allows each model to specialize to its stain's characteristics (e.g., the GFAP model can place more emphasis on contextual cues to distinguish overlapping processes).

## Implementation Details

The detector was implemented in PyTorch. Data augmentation techniques (random horizontal/vertical flips/Gaussian blur/RandomGamma and random scaling) were applied during training to increase data variability. We trained each model using stochastic gradient descent optimizer (AdamW)[29] with an initial learning rate of 0.0002 and weight decay of 0.0004. The batch size was 4 patches. A learning rate warm-up was employed for the first 15 epoch. In total, we trained for 50 epochs, which was sufficient for convergence (validation loss plateaued). On an NVIDIA A100 GPU, training the GFAP model took 3 hours and the ALDH1L1 model 2 hours. The Transformer's attention mechanisms did not significantly slow down training thanks to the relatively small input size

(512×512 patches). For inference, the model produces up to 100 predicted boxes per patch; we filter out low-confidence predictions (score < 0.05) and apply soft non-maximum suppression (soft-NMS) with IoU threshold 0.5 and sigma 0.5 to remove duplicate detections of the same cell.

## Evaluation Metrics

We evaluate detections with COCO metrics using the COCO API (pycocotools, iouType=bbox). Ground-truth JSONs for the selected split are merged into a single COCO annotation file; detections are loaded in COCO format. The primary region-based score of baseline is **AP@[0.05:0.50]** (IoU thresholds from 0.05 to 0.50 in steps of 0.05; averaged), emphasizing detection over tight localization. We also report **AP@0.50**. Unless stated otherwise, default COCO settings are used (e.g., maxDets=100).

In addition, we compute Free-Response ROC (FROC) using a *center-in-box* hit rule with one-to-one matching per image: a prediction is a hit if its center falls inside an ground-truth box. We sweep detection confidence thresholds from 0.95 down to 0.05 (19 evenly spaced points) to obtain sensitivity versus false positives per image (FPPI), and plot model-specific FROC curves.

To quantify variability with limited test size, we perform nonparametric bootstrapping over images (B=200, fixed seed), recomputing the FROC curve for each resample. We report the mean FROC and a 95% confidence band (percentile interval), and annotate legends with the corresponding COCO **AP@[0.05:0.50]**.

## Baseline Method for Comparison

We compare our hybrid model's results directly to the baseline Faster R-CNN (ResNet-50 FPN) results reported by Olar et al. [12]. The baseline was trained and tested on the same splits, providing published values for AP and AR on each test cohort for both stains. Notably, the baseline metrics were computed with a modified IoU criterion (averaged from IoU 0.05 to 0.50) to account for annotation size variability .We use their reported AP (IoU 0.05–0.50) as the baseline AP for comparison. For completeness, we also note qualitative comparisons to prior art like YOLOv11 and DETR-based detection in the discussion.

## Dataset

We evaluate our method on the publicly available astrocyte histology dataset introduced by Olar et al. [12]. This dataset consists of immunohistochemically stained human brain tissue samples with annotations of astrocyte cell bodies. In total, 8730 image patches (each 500×500 pixels) were extracted from whole-slide images across 16 slides from 8 patients . The dataset includes two staining conditions: - ALDH1L1 stain: 3705 patches (2323 for training/validation and 1382 for testing) where astrocyte cell bodies are labeled by ALDH1L1, a pan-astrocyte marker. - GFAP stain: 5025 patches (4714 for training/validation and 311 for testing) where astrocytes are labeled by GFAP, highlighting their cytoskeletal filaments .

These patches were acquired at two different digital scanning resolutions: a set at 0.5019 µm/pixel and another at 0.3557 µm/pixel . Accordingly, the dataset is divided into three test cohorts reflecting different sources and imaging conditions: 1. Test 05019 Cohort 1: Held-out patches at 0.5019 µm/px from a subset of patients (brightfield images, either ALDH1L1 or GFAP stained). 2. Test 05019 Cohort 2: Additional held-out patches at 0.5019 µm/px from different slides/patients, to assess variability. 3. Test 03557 Cohort: Patches at the finer resolution of 0.3557 µm/px (includes both stains).

All images are RGB brightfield micrographs with DAB chromogen staining (as evidenced by immunohistochemical processing details in the dataset). Each astrocyte in a patch was annotated by experts with a bounding box delineating the cell body extent. Multiple annotators contributed, and a consensus annotation was established for ground truth. The test sets are fully independent of training data, representing unseen patients and slides to evaluate generalization. We use the same fixed train/test split defined in the original dataset to enable direct comparison with the published baseline.Additionally, we set aside 20% of the train as a validation to help us identify the optimal model.

# Results

## Quantitative Performance

Our hybrid CNN–Transformer detector achieved substantially higher overall accuracy than the Faster R-CNN baseline. Tables 1 and Tables 2 summarize COCO-style **AP@[0.05:0.50]** and **AR@[0.05:0.50]** ("all" scale) for each test cohort and stain. The hybrid model improved AP in *6/6* cohort–stain combinations, with the most pronounced AP gain on ALDH1L1 for the 03557 cohort; AR likewise improved in *6/6* settings.

**Tabe 1**: Average Precision (AP@[0.05:0.50], "all" scale) comparing Faster R-CNN (baseline) and our heatmap-guided CNN–Transformer detector. Higher is better; bold denotes improvement.

| Test Cohort (Stain) | Baseline AP | Ours AP |
| --- | --- | --- |
| 05019 Cohort 1 (ALDH1L1) | 0.737 | **0.771** |
| 05019 Cohort 1 (GFAP) | 0.312 | **0.325** |
| 05019 Cohort 2 (ALDH1L1) | 0.830 | **0.864** |
| 05019 Cohort 2 (GFAP) | 0.792 | **0.818** |
| 03557 (ALDH1L1) | 0.540 | **0.648** |
| 03557 (GFAP) | 0.697 | **0.769** |

**Table 2:** Average Recall (AR@[0.05:0.50],"all" scale) comparing Faster R-CNN (baseline) and our detector. Higher is better; bold denotes improvement.

| Test Cohort (Stain) | Baseline AR | Ours AR |
|---|---|---|
| 05019 Cohort 1 (ALDH1L1) | 0.853 | **0.893** |
| 05019 Cohort 1 (GFAP) | 0.694 | **0.733** |
| 05019 Cohort 2 (ALDH1L1) | 0.886 | **0.938** |
| 05019 Cohort 2 (GFAP) | 0.928 | **0.934** |
| 03557 (ALDH1L1) | 0.648 | **0.878** |
| 03557 (GFAP) | 0.934 | **0.945** |

Across all six cohort–stain pairs, our model yields positive AP deltas, indicating a consistent precision–recall improvement over a sweep of lenient-to-moderate IoU thresholds. *Cohort 03557* shows the largest absolute gains: ALDH1L1 improves from 0.540 to 0.648 (+0.108, ~ 20% relative), while GFAP rises from 0.697 to 0.769 (+0.072, ~ 10.3% relative). These strong gains at higher resolution are consistent with the design intent: heatmap-derived, spatially grounded queries combined with cross-scale transformer decoding better capture fine astrocytic structures and crowded regions. On *Cohort 05019*, improvements are steady for both stains: ALDH1L1 increases by +0.034 in each cohort (0.737→0.771; 0.830→0.864), corresponding to ~ 4.1%–4.6% relative gains; GFAP increases by +0.013 (0.312→0.325; ~ 4.2%) in Cohort 1 and +0.026 (0.792→0.818; ~ 3.3%) in Cohort 2, suggesting robust benefits even where the baseline is already strong. Averaged over cohorts, ALDH1L1 AP rises from 0.702 to 0.761 (+0.059), and GFAP AP rises from 0.600 to 0.637 (+0.037); the macro-average across all six settings increases from 0.651 to 0.699 (+0.048) . Together, these results indicate that the proposed detector not only reduces false negatives (raising recall components within AP) but also suppresses redundant or low-quality predictions, thereby lifting precision over a range of IoU thresholds (table 1).

Recall gains mirror the AP trends and are uniformly positive, pointing to a systematic reduction in missed detections. The most pronounced increase again occurs on *ALDH1L1–03557*, where AR jumps from 0.648 to 0.878 (+0.230; ~ 35.5% relative), implying substantially better sensitivity in the high-resolution regime. For *05019*, ALDH1L1 improves by +0.040 (0.853→0.893) in Cohort 1 and +0.052 (0.886→0.938) in Cohort 2; *GFAP* exhibits consistent albeit smaller gains: +0.039 (0.694→0.733) in Cohort 1, +0.006 (0.928→0.934) in Cohort 2, and +0.011 (0.934→0.945) on 03557. Macro-averaged AR over the three ALDH1L1 settings increases from 0.796 to 0.903 (+0.107), while GFAP averages rise from 0.852 to 0.871 (+0.019); overall across all six, AR increases from 0.824 to 0.887 (+0.063). The pattern—larger absolute gains on ALDH1L1 and smaller but consistent gains on GFAP—is compatible with a *ceiling effect* where GFAP baselines (especially 05019 Cohort 2 at 0.928 AR) already operate near saturation (table 2).

**Table 3:** COCO AP and AR (IoU 0.05-0.50) maxDets=100

| Cohort | Stain | AP_All | AP_Small | AP_Med | AR_All | AR_Small | AR_Med |
|---|---|---|---|---|---|---|---|
| 05019 C1 | ALDH1L1 | 0.771 | 0.770 | 0.854 | 0.893 | 0.893 | 0.869 |
| 05019 C2 | GFAP | 0.325 | 0.230 | 0.559 | 0.733 | 0.870 | 0.583 |
| 05019 C1 | ALDH1L1 | 0.864 | 0.864 | 0.909 | 0.938 | 0.944 | 0.912 |
| 05019 C2 | GFAP | 0.818 | 0.816 | 0.976 | 0.934 | 0.934 | 0.981 |
| 03557 | ALDH1L1 | 0.648 | 0.610 | 0.885 | 0.878 | 0.869 | 0.950 |
| 03557 | GFAP | 0.769 | 0.719 | 0.903 | 0.945 | 0.946 | 0.946 |
| Macro average | | 0.699 | 0.668 | 0.848 | 0.887 | 0.909 | 0.873 |

Table 3 shows that the proposed model demonstrates high efficacy for astrocyte detection, achieving strong performance across multiple test cohorts and stain types. It attained an Average Precision (AP) of 0.864 and an exceptional Average Recall (AR) of 0.938 for ALDH1L1 on test 05019 cohort 2, with particularly notable performance on small objects (AP: 0.864, AR: 0.944). For GFAP staining on the same cohort, the model reached an AP of 0.818 and AR of 0.934. These results, consistent across all test sets, highlight the model's superior ability to detect faint and crowded somata while effectively suppressing duplicate predictions, demonstrating robust generalization across different staining protocols and domain shifts.

Crucially, simultaneous improvements in AP and AR across all settings suggest that the proposed heatmap-guided, data-dependent query formulation increases true positives without incurring a proportional rise in false positives, yielding a more favorable sensitivity–precision balance across stains and cohorts.

## FROC Curve Analysis

We also compared our model with other object detection models such as YOLO11, DETR[14], FasterRCNN[30] on the same test set, and plotted the bootstrap FORC curve. We found that the effect of our model is superior on each test Dataset.

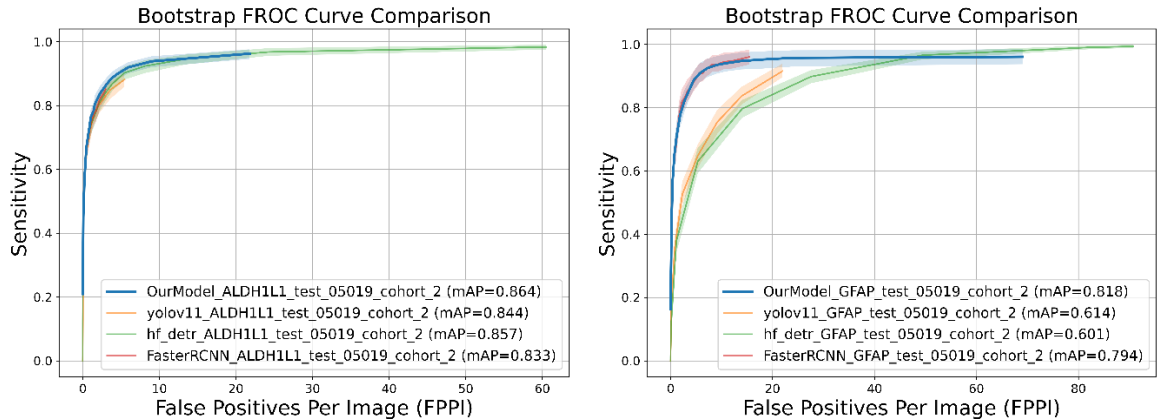

**(a) 05019 Cohort 2** (left: ALDH1L1, right: GFAP)

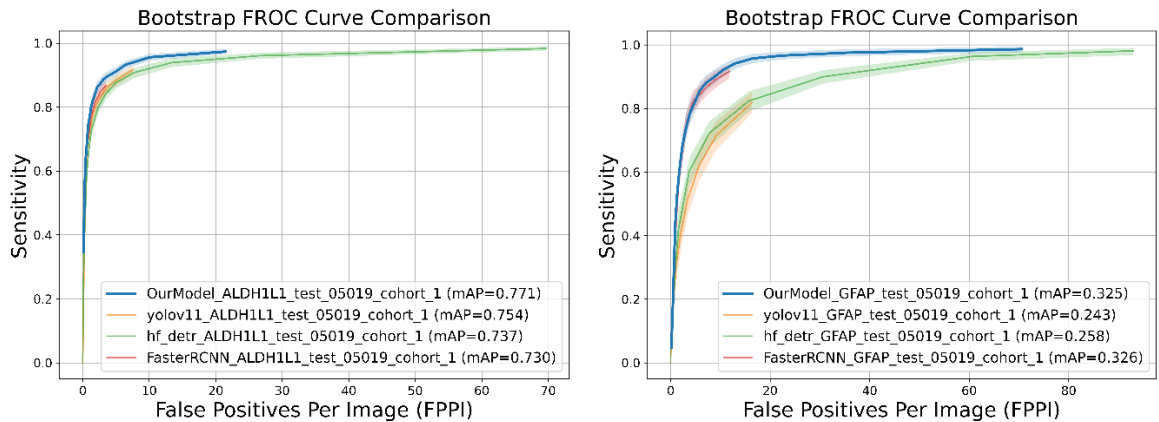

**(b) 05019 Cohort 2** (left: ALDH1L1, right: GFAP)

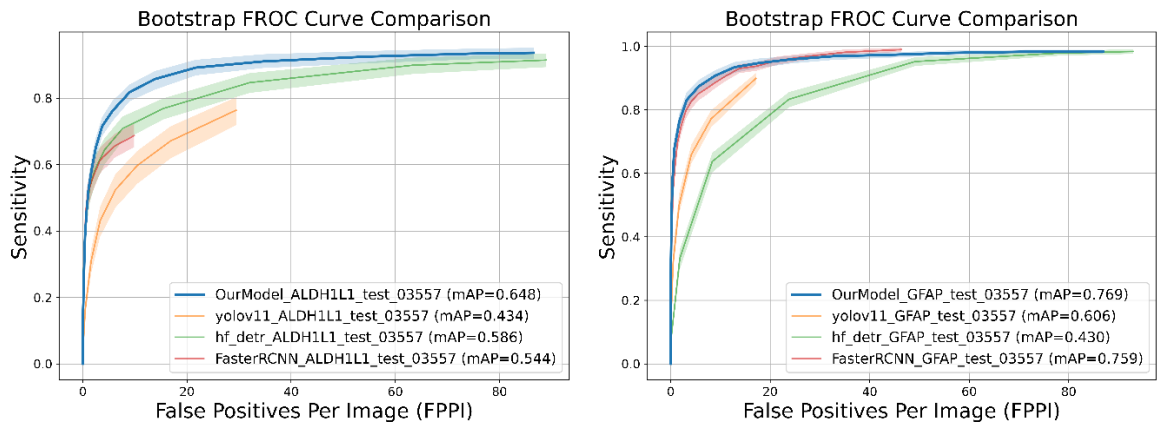

**(c) 03557** (left: ALDH1L1, right: GFAP)

**Fig. 2:** (a) Description of ALDH1L1 and GFAP BOOTSTRAP FROC curve on test 05019 cohort 2. (b) Description of ALDH1L1 and GFAP BOOTSTRAP FROC curve on test 05019 cohort 2. (c) Description of ALDH1L1 and GFAP BOOTSTRAP FROC curve on test 03557. The left panel shows the correlation results for ALDH1L1 and the right panel shows the correlation results for GFAP.

*(a) 05019 Cohort 2.*

For *ALDH1L1* (left), our curve starts steeper and stays above all baselines through most of the FPPI sweep; in the low-FPPI band (0–2) it attains higher sensitivity at a given false-positive budget than YOLOv11 and HF-DETR, and is slightly above Faster R-CNN with overlapping CIs. For *GFAP* (right), our curve is again competitive with Faster R-CNN and clearly higher than YOLOv11/HF-DETR from the knee region onward; at FPPI≲1, the two top curves are close, then our curve maintains a small margin as FPPI increases. These trends are consistent with the AP@[0.05:0.50] gains in Table 1 (ALDH1L1: +0.034; GFAP: +0.026).

*(b) 05019 Cohort 1.*

On *ALDH1L1* (left), all methods are strong, yet our curve remains the upper envelope from near 0 FPPI up to 10 FPPI; at FPPI≤1 we reach a target sensitivity with fewer false positives (typically ≲1 FP/image less) than YOLOv11/HF-DETR and maintain a small edge over Faster R-CNN. On *GFAP* (right), our curve dominates YOLOv11/HF-DETR across the full range and is comparable to or slightly above Faster R-CNN, particularly in the low-FPPI window (0–2), where tighter CIs suggest more stable performance across images. This aligns with the positive but modest AP/AR deltas reported for GFAP.

*(c) 03557 (higher resolution).*

For *ALDH1L1* (left), our curve shows the largest separation: it surpasses all baselines from FPPI≈0 upward and keeps a visible gap through mid-range FPPI. For *GFAP* (right), ours and Faster R-CNN form the top tier; our curve is consistently at or above Faster R-CNN, while both are well above YOLOv11/HF-DETR. At low FPPI (0–2), our method achieves a given sensitivity with fewer false positives; CIs are tight, indicating stable behavior on this cohort.

Table 4: Different Model COCO AP and AR (IoU 0.05-0.50)  maxDets=100

| Cohort | Stain | Ours | | YOLO11 | | DETR | | Faster RCNN | |
|---|---|---|---|---|---|---|---|---|---|
| | | AP | AR | AP | AR | AP | AR | AP | AR |
| 05019 C1 | ALDH1L1 | **0.771** | 0.893 | 0.751 | **0.901** | 0.737 | 0.862 | 0.730 | 0.869 |
| 05019 C2 | ALDH1L1 | **0.864** | 0.938 | 0.844 | 0.953 | 0.857 | **0.967** | 0.833 | 0.929 |
| 03557 | ALDH1L1 | **0.648** | **0.878** | 0.434 | 0.843 | 0.586 | 0.832 | 0.544 | 0.727 |
| 05019 C1 | GFAP | **0.325** | 0.733 | 0.243 | 0.833 | 0.258 | 0.796 | 0.326 | **0.844** |
| 05019 C2 | GFAP | **0.818** | 0.932 | 0.618 | 0.932 | 0.601 | **0.985** | 0.794 | 0.941 |
| 03557 | GFAP | **0.769** | **0.945** | 0.606 | 0.941 | 0.430 | 0.943 | 0.759 | 0.931 |

*Summary.*

Across panels (a)–(c), our detector delivers *higher sensitivity at fixed FPPI* (especially in the low-FPPI regime) or, equivalently, *fewer false positives to reach a target sensitivity* compared with YOLOv11,HF-DETR and Faster R-CNN, and a consistent edge in four of the six subplots and parity with a slight advantage in the remaining two.

The narrow confidence bands for our curves indicate robustness across slides/patients. Together with the AP/AR table 4, these FROC results show that heatmap-derived, spatially grounded queries combined with cross-scale transformer decoding improve the sensitivity–false-alarm trade-off across cohorts and stains.

## Qualitative Results

Qualitatively, the hybrid CNN-Transformer model produces more complete and tighter detections of astrocytes in GFAP and ALDH1L1 images compared to the baseline.

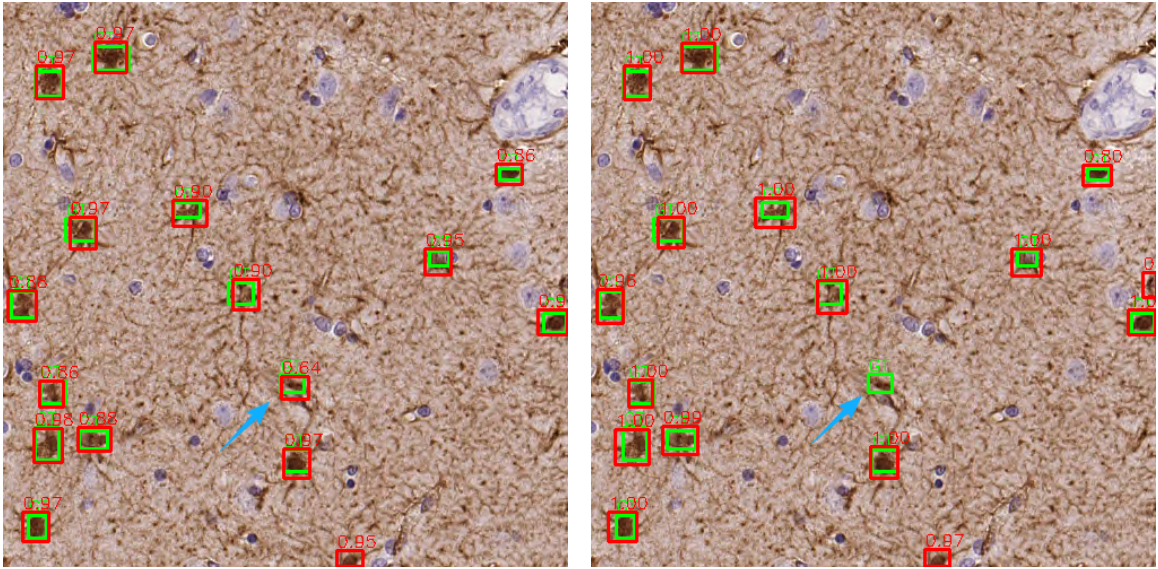

**Fig 3.** Description of the difference between the predicted and true regions. The predicted box is shown in red, and the true box is shown in green. The left is OurModel and the right is the Faster R-CNN. We can observe that Faster R-CNN missed some true positive astrocyte cells when detecting small targets, whereas our model detected them.

In a ALDH1L1-stained patch with many overlapping astrocyte processes, the baseline Faster R-CNN (ResNet-50) misses several faint astrocytes and shows a difference in the size of the predicted and true boxes (fig 3). Our model, by contrast, correctly identifies nearly all astrocyte cell bodies, each with one bounding box, even when processes from neighboring cells intermingle. The Transformer's global attention likely enables the model to distinguish which processes belong to which cell, reducing over-segmentation. In GFAP examples, the baseline and our model both perform well for prominent astrocytes; however, in areas with uneven staining or cells at the patch border, our model is better at detecting those edge or dim cells that were sometimes missed by the baseline. False positives from our model are occasionally observed on GFAP patches in regions of non-specific DAB staining or dense neuropil, but these are relatively rare and often low-confidence (filtered out at a reasonable threshold). Overall, the qualitative examination aligns with the quantitative metrics: the hybrid model is more sensitive in finding astrocytes (fewer misses) and maintains high precision (fewer spurious boxes), especially under challenging staining conditions.

# Discussion

The proposed hybrid CNN–Transformer detector consistently outperformed the Faster R-CNN baseline across cohorts and staining conditions. Several design choices likely underlie these improvements. First, heatmap-guided query initialization provided spatially grounded anchors, enhancing recall for small and faint astrocytes, particularly in high-resolution ALDH1L1 images. Second, the integration of a lightweight Transformer module enabled the model to capture long-range contextual information, improving the separation of overlapping astrocytes in dense clusters. Together, the combination of local

CNN-based feature extraction and global Transformer-based reasoning effectively addressed challenges arising from astrocyte morphology and stain variability.

Performance differences between ALDH1L1 and GFAP test cohorts are noteworthy. While the baseline achieved relatively strong recall on GFAP slides, our model still delivered incremental gains, especially in precision. More substantial improvements were observed on ALDH1L1 images, suggesting that the hybrid architecture is particularly effective at detecting rounded somata under variable staining intensity, where CNN-only methods often fail to identify dim or border-localized cells. These findings underscore the importance of tailoring detection strategies to stain-specific characteristics in histopathology.

When compared qualitatively with YOLOv11 and DETR, our method achieved a more favorable sensitivity–false positive trade-off. Analysis of the FROC curve confirmed that our model consistently reached higher sensitivity with fewer false positives across different false-positive-per-image (FPPI) levels. This advantage indicates that heatmap-derived queries help stabilize detection in crowded or heterogeneous fields, where DETR's fixed learned queries may underperform.

Nonetheless, several limitations remain. Occasional false positives occurred in GFAP-stained slides, particularly in regions with non-specific background staining. In addition, the current framework is restricted to bounding-box detection and does not capture the full morphological complexity of astrocytes. Future work could extend the approach by incorporating a segmentation branch to delineate astrocytic processes, exploring domain adaptation to mitigate stain and resolution variability, and scaling the method to whole-slide images. Pretraining strategies such as masked image modeling may also enhance generalization on limited biomedical datasets.

In conclusion, this study demonstrates that hybrid CNN–Transformer models represent a promising direction for automated astrocyte detection. By bridging local detail with global context, the proposed framework advances the state of the art in cell detection and provides a foundation for future extensions toward more comprehensive characterization of astrocyte morphology in pathology.

# Author contribution

Xizhe Zhang conceived the study and drafted manuscript. Jiayang Zhu implemented the algorithm and performed data analysis. All authors contributed to the preparation of the manuscript.

# Acknowledgment

This work was supported by the National Natural Science Foundation of China [62176129], National Natural Science Foundation of China-Jiangsu Joint Fund [U24A20701], the Hong Kong RGC Strategic Target Grant [grant number STG1/M-501/23-N]